\documentclass[preprint]{ccn}

\title{Heterogeneous Time Constants Improve Stability in Equilibrium Propagation}

\author{%
  Yoshimasa Kubo\affmark{1} \And
  Suhani Pragnesh Modi\affmark{1} \And
  Smit Patel\affmark{1} \AND
}
\affiliation{1}{Department of Computer Science, Lakehead University}
\emails{ykubo@lakeheadu.ca, smodi7@lakeheadu.ca, spate239@lakeheadu.ca}

\begin{document}

\maketitle

\begin{abstract}
Equilibrium propagation (EP) is a biologically plausible alternative to backpropagation for training neural networks. However, existing EP models use a uniform scalar time step 
$dt$, which corresponds biologically to a membrane time constant that is heterogeneous across neurons. Here, we introduce heterogeneous time steps (HTS) for EP by assigning neuron-specific time constants drawn from biologically motivated distributions. We show that HTS improves training stability while maintaining competitive task performance. These results suggest that incorporating heterogeneous temporal dynamics enhances both the biological realism and robustness of equilibrium propagation.
\end{abstract}

\section{Introduction}
Equilibrium propagation (EP) is a biologically plausible learning algorithm for training neural networks \citep{Scellier2017,Scellier2019,Ernoult2019,Laborieux2021,Laborieux2022}. Prior work has applied EP to image classification \citep{Laborieux2021,Laborieux2022}, including continual learning \citep{kubo2025}, reinforcement learning \citep{kubo2022b}, and natural language processing tasks, highlighting its potential as an alternative to backpropagation, which, despite its effectiveness, lacks biological plausibility. Nevertheless, several aspects of EP remain underexplored from a biological perspective. In particular, existing EP implementations typically employ a single scalar time step $dt$ to update neural states. Interpreted biologically, this parameter corresponds to a membrane time constant, which is known to vary substantially across neurons rather than remaining uniform. Moreover, prior studies in spiking neural networks have shown that heterogeneous time constants can improve performance on multiple tasks \citep{perez2021}.

Here, we introduce heterogeneous time steps (HTS) for equilibrium propagation, assigning neuron-specific time constants drawn from biologically motivated distributions instead of using a shared scalar value. We evaluate multiple distributions and analyze their effects on learning dynamics and performance. Our results show that EP models with HTS exhibit improved training stability compared to conventional scalar-time-step EP, while maintaining competitive accuracy. These findings suggest that incorporating heterogeneous temporal dynamics enhances both the biological plausibility and robustness of equilibrium propagation.

\begin{figure*}[h!t]
    \centering
    \includegraphics[height=5.0cm,width=1.0\textwidth]{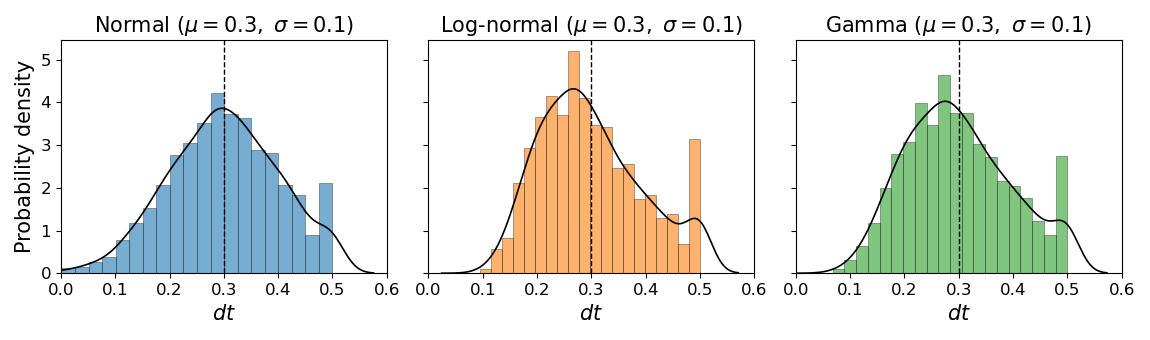}
    \caption{Distributions of heterogeneous time steps ($dt_i$) used in the hidden layer.
All distributions were parameterized with $\mu=0.3$, $\sigma=0.1$ and clamped to $[10^{-3}, 0.5]$ for numerical stability. The visible accumulation at $dt=0.5$ for heavy-tailed distributions results from this truncation. }
    \label{fig:dt_distributions}
\end{figure*}

\section{Methods}
\subsection{Model Dynamics}

Equilibrium propagation (EP) formulates neural state dynamics as gradient flow on an energy function $E$. In continuous time, the dynamics are expressed as
\begin{equation}
\tau \frac{ds}{dt} = - \frac{\partial E}{\partial s},
\end{equation}
where $\tau$ denotes the neuronal membrane time constant and $s$ represents the neural state.

In discrete-time implementations, this yields the update rule
\begin{equation}
s^{t+1} = s^{t} - dt \, \frac{\partial E}{\partial s^{t}},
\end{equation}
where $dt$ is the integration time step. In conventional EP models, $dt$ is treated as a scalar shared across all neurons.

For state updates, we employed the predictive learning rule \citep{luczak2022a} without explicit prediction units \citep{luczak2022b,kubo2022a}, resulting in a more biologically grounded implementation.

\begin{table*}[t]
\centering
\caption{Average test accuracy (\%) $\pm$ standard deviation across 10 random seeds.}
\begin{tabular}{|c|c|c|c|c|c|}
\hline
\textbf{Dataset} & \textbf{$dt_y$} & \textbf{Normal} & \textbf{Log-normal} & \textbf{Gamma} & \textbf{Scalar} \\
\hline
\textbf{MNIST} & 0.15 & 98.43 $\pm$ 0.03\% & 98.43 $\pm$ 0.04\% & 98.43 $\pm$ 0.04\% & 98.43 $\pm$ 0.04\% \\
& 0.20 & 98.44 $\pm$ 0.03\% & 98.44 $\pm$ 0.02\% & 98.43 $\pm$ 0.02\% & 98.43 $\pm$ 0.04\% \\
& 0.25 & 98.44 $\pm$ 0.04\% & 98.44 $\pm$ 0.03\% & 98.43 $\pm$ 0.03\% & 98.44 $\pm$ 0.04\% \\
& 0.30 & 98.46 $\pm$ 0.03\% & 98.46 $\pm$ 0.02\% & 98.45 $\pm$ 0.04\% & 98.43 $\pm$ 0.03\% \\
& 0.35 & 98.46 $\pm$ 0.04\% & 98.45 $\pm$ 0.03\% & 98.45 $\pm$ 0.04\% & 98.44 $\pm$ 0.04\% \\
\hline
\textbf{KMNIST}  & 0.15 & 91.29 $\pm$ 0.22\% & 91.21 $\pm$ 0.14\% & 91.15 $\pm$ 0.18\% & 91.07 $\pm$ 0.17\% \\
& 0.20 & 91.21 $\pm$ 0.17\% & 91.22 $\pm$ 0.13\% & 91.23 $\pm$ 0.15\% & 91.10 $\pm$ 0.18\% \\
& 0.25 & 91.23 $\pm$ 0.11\% & 91.24 $\pm$ 0.11\% & 91.28 $\pm$ 0.11\% & 91.11 $\pm$ 0.07\% \\
& 0.30 & 91.26 $\pm$ 0.14\% & 91.19 $\pm$ 0.11\% & 91.26 $\pm$ 0.15\% & 91.12 $\pm$ 0.15\% \\
& 0.35 & 91.23 $\pm$ 0.16\% & 91.27 $\pm$ 0.16\% & 91.17 $\pm$ 0.15\% & 91.10 $\pm$ 0.18\% \\
\hline
\textbf{FMNIST}  & 0.15 & 89.65 $\pm$ 0.13\% & 89.65 $\pm$ 0.13\% & 89.65 $\pm$ 0.11\% & 89.53 $\pm$ 0.12\% \\
& 0.20 & 89.69 $\pm$ 0.16\% & 89.71 $\pm$ 0.10\% & 89.66 $\pm$ 0.13\% & 89.44 $\pm$ 0.17\% \\
& 0.25 & 89.65 $\pm$ 0.10\% & 89.66 $\pm$ 0.11\% & 89.63 $\pm$ 0.18\% & 89.57 $\pm$ 0.23\% \\
& 0.30 & 89.68 $\pm$ 0.18\% & 89.63 $\pm$ 0.08\% & 89.61 $\pm$ 0.13\% & 89.50 $\pm$ 0.16\% \\
& 0.35 & 89.65 $\pm$ 0.12\% & 89.59 $\pm$ 0.12\% & 89.64 $\pm$ 0.16\% & 89.43 $\pm$ 0.23\% \\
\hline
\end{tabular}
\label{tbl:results_perform}
\end{table*}

\subsection{Heterogeneous Time Steps}

To introduce temporal heterogeneity in EP, we replaced the conventional scalar $dt$ with neuron-specific time steps in the hidden layer. Each hidden neuron was assigned a distinct $dt_i$ sampled from a probability distribution. We evaluated three biologically motivated distributions: normal, log-normal, and gamma.

All distributions were parameterized with mean $0.3$ and standard deviation $0.1$. To ensure numerical stability of the discrete-time Euler updates and to avoid pathological time scales, all sampled time steps were restricted to a bounded interval:
\[
dt_i \in [dt_{\min}, dt_{\max}],
\]
where $dt_{\min} = 10^{-3}$ and $dt_{\max} = 0.5$.
Bounding $dt_i$ prevents excessively large integration steps that could destabilize gradient-flow dynamics, as well as vanishingly small values that would significantly slow convergence. 

For heavy-tailed distributions such as the log-normal and gamma, samples exceeding $dt_{\max}$ are mapped to $dt_{\max}$, which results in a small accumulation of probability mass at the upper boundary. This effect is visible in Figure~\ref{fig:dt_distributions} and reflects the imposed stability constraint rather than an intrinsic property of the underlying distributions.

Heterogeneous time steps were applied only to the hidden layer, while the output layer retained a scalar time step due to its distinct functional role in classification. The hidden layer contained 1024 neurons in all experiments.

To examine the interaction between hidden-layer heterogeneity and output-layer dynamics, we varied the scalar time step of the output layer across $\{0.15, 0.20, 0.25, 0.30, 0.35\}$. This allowed us to evaluate stability and performance under different output temporal scales while maintaining heterogeneous dynamics in the hidden layer.

\subsection{Experimental Setup}
Experiments were conducted on MNIST, KMNIST, and Fashion-MNIST using a single hidden layer with 1024 neurons. The hidden and output activation functions were Leaky ReLU and sigmoid, respectively. Models were trained for 50 epochs with batch size 256. Learning rates were set to $\alpha_1 = 0.5$ (input-to-hidden) and $\alpha_2 = 0.1$ (hidden-to-output). The free and clamped phases consisted of 125 and 12 time steps, respectively, with feedback parameter $\gamma = 1.0$ and nudging parameter $\beta = 1.0$.

\section{Results and Conclusion}

Table~\ref{tbl:results_perform} summarizes test accuracy across datasets. On MNIST, performance differences among models are negligible. In contrast, on KMNIST and Fashion-MNIST, EP models with heterogeneous time steps (HTS) show consistent but modest improvements over the scalar baseline, suggesting a mild regularizing effect similar to activation-based regularization in EP \citep{kubo2022a}.

In summary, incorporating heterogeneous time steps improves stability while preserving performance and increasing biological plausibility.

\section*{Acknowledgments}

This research was enabled in part by computational resources provided by the Digital Research Alliance of Canada (alliancecan.ca).

\nocite{Scellier2017}
\nocite{Scellier2019}
\nocite{Ernoult2019}
\nocite{Laborieux2021}
\nocite{Laborieux2022}
\nocite{kubo2022b}
\nocite{kubo2025}
\nocite{perez2021}
\nocite{kubo2022a}
\nocite{luczak2022a}
\nocite{luczak2022b}
\printbibliography

\end{document}